\documentclass[letterpaper, 10 pt, conference]{ieeeconf}  % Comment this line out if you need a4paper

\IEEEoverridecommandlockouts                              % This command is only needed if 
\usepackage{graphicx}
\usepackage{setspace}
\usepackage{mathptmx}
\usepackage{bm}
\usepackage{amsmath,amssymb} % define this before the 
\usepackage{color}
\usepackage{caption}
\usepackage{algorithm}
\usepackage{array}
\newcolumntype{C}[1]{>{\centering\let\newline\\\arraybackslash\hspace{0pt}}m{#1}}
\usepackage{algpseudocode}
\usepackage{enumerate}
\usepackage{epstopdf}
\usepackage{array}
\usepackage{eqparbox}
\usepackage{multirow}
\usepackage{url}
\usepackage{mathtools}
\usepackage{hyperref}
\usepackage{float}
\usepackage[section]{placeins}
\title{\LARGE \bf
Object Parsing in Sequences Using CoordConv Gated Recurrent Networks
}

\author{Ayush Gaud$^{1}$, Y V S Harish$^{1}$ and K Madhava Krishna$^{1}$% <-this % stops a space
\thanks{$^{1}$Ayush Gaud, Y V S Harish and K Madhava Krishna are with Robotics Research Center, International Institute of
Information Technology, Hyderabad, India.
        {\tt\small ayush.gaud@gmail.com harish.y@research.iiit.ac.in
        mkrishna@iiit.ac.in}}%
}

\begin{document}
\makeatletter
\let\@oldmaketitle\@maketitle
\renewcommand{\@maketitle}{\@oldmaketitle
\centering
\includegraphics[width=17cm, height=5.5cm] {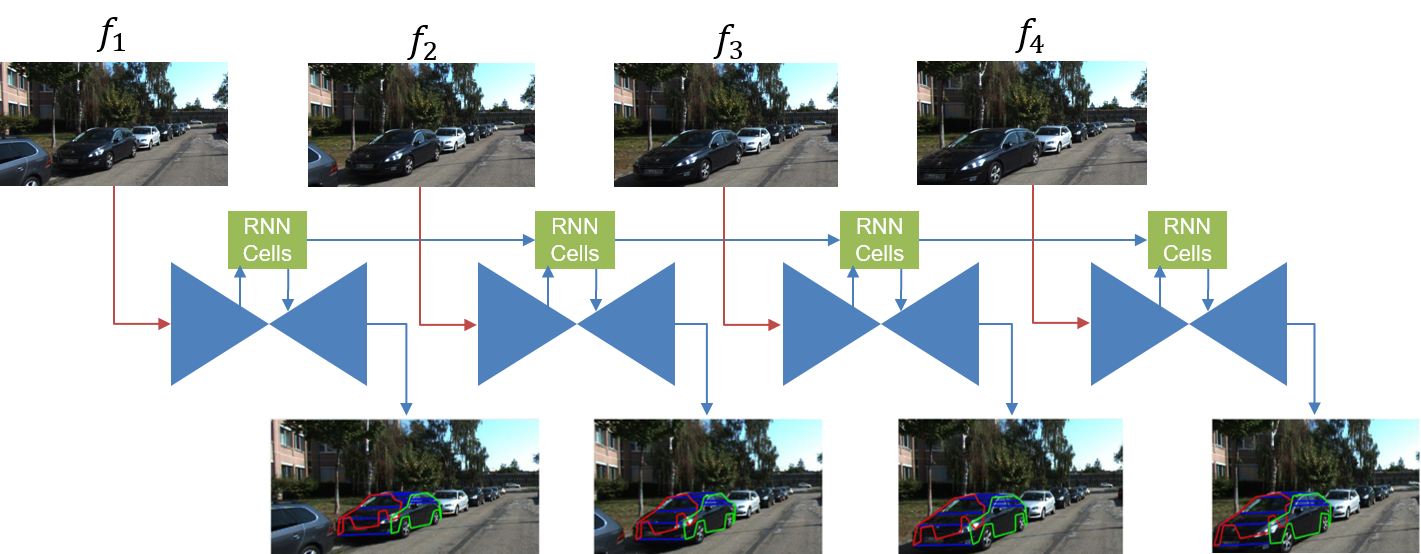}
    \captionof{figure}{Overview of our approach. We propose a recurrent hourglass network based on CoordConvGRU cells. We demonstrate that the spatio-temporal consistency is preserved and we generate increasingly refined estimates over sequential data. All the network modules share the same weight across stages while the hidden states from the recurrent cell is passed as input for the next iteration of keypoint localization on subsequent frames.}
}
\label{fig:overall_pipeline}
\makeatother

\maketitle
%\thispagestyle{empty}
%\pagestyle{empty}
%%%%%%%%%%%%%%%%%%%%%%%%%%%%%%%%%%%%%%%%%%%%%%%%%%%%%%%%%%%%%%%%%%%%%%%%%%%%%%%%
\begin{abstract}
We present a monocular object parsing framework for consistent keypoint localization by capturing temporal correlation on sequential data. In this paper, we propose a novel recurrent network based architecture to model long-range dependencies between intermediate features which are highly useful in tasks like keypoint localization and tracking. We leverage the expressiveness of the popular stacked hourglass architecture and augment it by adopting memory units between intermediate layers of the network with weights shared across stages for video frames. We observe that this weight sharing scheme not only enables us to frame hourglass architecture as a recurrent network but also prove to be highly effective in producing increasingly refined estimates for sequential tasks. Furthermore, we propose a new memory cell, we call CoordConvGRU which learns to selectively preserve spatio-temporal correlation and showcase our results on the keypoint localization task. The experiments show that our approach is able to model the motion dynamics between the frames and significantly outperforms the baseline hourglass network. Even though our network is trained on a synthetically rendered dataset, we observe that with minimal fine tuning on 300 real images we are able to achieve performance at par with various state-of-the-art methods trained with the same level of supervisory inputs. By using a simpler architecture than other methods enables us to run it in real time on a standard GPU which is desirable for such applications. Finally, we make our architectures and 524 annotated sequences of cars from KITII dataset publicly available.
\end{abstract}

%%%%%%%%%%%%%%%%%%%%%%%%%%%%%%%%%%%%%%%%%%%%%%%%%%%%%%%%%%%%%%%%%%%%%%%%%%%%%%%%

\section{INTRODUCTION}
% \begin{figure*}[!ht]
%     \begin{center}
%     \includegraphics[width=16cm, height=5.5cm] {figures/Pipeline.png}
%     \caption{Overview of our approach. We propose a recurrent hourglass network based on CoordConvGRU cells. We demonstrate that the spatio-temporal consistency is preserved and we generate increasingly refined estimates over sequential data. All the network modules shares the same weight across stages while the hidden states from the recurrent cell is passed as input for the next iteration of keypoint localization on subsequent frames.}
%     \label{fig:overall_pipeline}
%     \end{center}
% \end{figure*}
Estimating semantically meaningful joint locations and hence shape and pose of moving objects like cars in a highway driving scenario is a challenging problem in computer vision. With the growing interest in autonomous driving, many successful approaches use expensive sensors like LiDARs and stereo cameras for tracking and predicting trajectory of vehicles on road. However, there is an ongoing effort to minimize the cost of expensive sensor suite required for such tasks. Some recent approaches have shown to recover both shape and pose from monocular images but fail to leverage the temporal characteristics of the data in such scenarios\cite{murthy2017shape}. Our approach bases itself on the success of the stacked hourglass architecture\cite{newell2016stacked} which showcased state of the art results for human pose prediction. While its multi-level encoder-decoder architecture with skip layers preserves spatial information at different scales, it fails to exploit any temporal information that is present in sequential data and treats each frame independently which is detrimental especially in scenarios like driving.

In this paper, we posit that leveraging the temporal information using recurrent architectures enables us to parse objects with higher accuracy. It is based on the notion that the corresponding part locations \emph{(keypoints)} of an object is usually in the neighborhood across frames of a sequence, especially for a rigid body. We present our findings which conforms to this hypothesis and justifies the validity of our proposed network architecture for performing this task. Specifically, we demonstrate that on scenarios like driving where environment involves high dynamics, this property can be quite useful for tasks like propagating annotations on on-road vehicles consistently across the frames, unlike other methods\cite{murthy2017shape}\cite{newell2016stacked}\cite{kanazawa2016warpnet}\cite{li2017deep} which discards the sequential consistency of a scene by treating each frame independently to perform predictions.

To illustrate this idea, we utilize semantically meaningful locations of an object represented as keypoints which efficiently encapsulates the 3D structure of an object category. These 3D keypoints are annotated on CAD models which enables us to generate large amounts of synthetic dataset required for training an architecture like this. The dataset is generated using the annotated CAD models of cars from the ShapeNet\cite{chang2015shapenet} corpus. The cars are rendered from different viewpoints by moving the camera in a natural motion for generating sequences which appear temporally consistent. We observe a consistent improvement in the accuracy of predicted keypoint locations as we continue to improve the expressiveness of the model by making changes to the original hourglass architecture by adding ConvGRU and then CoordConvGRU cells to the residual layer. Our contributions are summarized as follows:
\begin{enumerate}
\item We introduce an improved stacked hourglass architecture to impose spatio-temporal consistency in consecutive video frames from a monocular camera for keypoint prediction. Instead of employing RNNs based on vanilla LSTM, we utilize convolutional gated recurrent networks\cite{ballas2015delving} part to preserve spatial connectivity in the image. 
\item We note further improvement using our proposed CoordConvGRU cells designed by adding two coordinate channels, one for each axis, to the convolution layer of ConvGRU, which learns to selectively preserve translational invariance and effectively learn the temporal correlation between intermediate features of the frames in a sequence.
% \subsubsection{}We further propose a differentiable bilinear resampling based supervision which enables network to develop improved temporal co-relation between intermediate frames of a sequence.
\end{enumerate}

We evaluate our hypothesis by conducting ablation experiments on different network designs to highlight the efficacy of our approach. To further emphasize the generalizability of our approach, we also perform benchmarks on the images from KITTI dataset and compare the predicted keypoint locations with various other networks. Our architecture is able to outperform several other architectures, while at the same time is fast enough to run in real-time on a standard GPU. In section 4 we present the qualitative and quantitative results of our proposed architecture and demonstrate consistent improvement in performance of the network across other architectures.
\section{RELATED WORK}
Keypoint localization is a well studied topic in the literature popularized by pose estimation for humans. DeepPose\cite{toshev2014deeppose} shifted the attention of people from classical approaches to deep learning based approaches. Tompson\cite{tompson2014joint} further brought a significant improvement in keypoint localization by introducing a fully convolution based network. He proposed that generating heatmaps directly from a hierarchy of multi-scale convolutional structure combined with a graphical model enables learning the spatial relationships between the joints and achieves higher accuracy. Tompson built his work\cite{tompson2014joint} on the idea of cascade refinement based on the multi-stage pose machines\cite{ramakrishna2014pose}.

Newell et al.\cite{newell2016stacked} introduced intermediate supervision to a cascaded network based on the conv-deconv and encoder-decoder architectures\cite{noh2015learning}\cite{zhao2015stacked}\cite{rematas2016deep}\cite{badrinarayanan2017segnet}. The repeated bottom-up and top-down inference from stacking further allows for reevaluation of initial estimates and features across the image. This helps in forming higher order spatial relationships and maintaining local and global cues.

In a few studies, temporal cues have also been integrated\cite{gkioxari2016chained}\cite{jain2014modeep}\cite{lin2017recurrent}\cite{xiaohan2015joint}\cite{pfister2015flowing}\cite{song2017thin} to the network for the task of pose estimation. Using dense optical flow\cite{weinzaepfel2013deepflow} certain approaches\cite{pfister2015flowing}\cite{song2017thin} have attempted in predicting consistent position of joints preserving smooth motion across the frames. Thin-Slicing network\cite{song2017thin} presented improved results by using both optical flow and a spatial-temporal model but at the cost of computational complexity rendering it slower than other approaches. 

LSTM pose machines\cite{luo2018lstm} introduced using of memory augmented recurrent networks to capture temporal consistency and was able to output perform various state of the art approaches in 2D video pose estimation tasks. Their design was based on another strong baseline approach Convolution Pose Machines\cite{wei2016convolutional} by using essentially the same architecture but also providing a weight sharing scheme across stages while at the same time using LSTM cells to promote intermediate features. This helps in better utilizing historical joint locations and achieving almost double inference speed which is critical for real-time applications.

Siam et al.\cite{siam2017convolutional} investigated convolutional gated recurrent architecture (convGRU) for video segmentation. They proposed a modification to the standard convolutional recurrent units which are designed for processing text data, to preserve the spatial connectivity between the pixels which would have been otherwise lost. By replacing the dot products with convolution operation they were able to create a much more efficient recurrent cell which can be easily trained on images without the curse of high dimensionality of weights. They were able to achieve significant gains on the baseline architectures by exploiting the spatio-temporal information of the videos.

Payer et al.\cite{payer2018instance}, further utilizing the effectiveness of both hourglass and ConvGRU architectures, presented a recurrent hourglass architecture for instance segmentation tracking. They rigorously evaluated their approach on various cell tracking datasets and were able to achieve state-of-the-art performance.

CoordConv layers presented by Liu et al.\cite{liu2018intriguing} posed a counter intuitive idea from assuming the convolution layers to be appropriate for spatial representations. This is justified by a trivial counterexample of learning coordinate transforms in one-hot pixel space. They claim that by adding a few extra coordinate channels enables the network to learn varying degrees of translational invariance. There is a stark contrast observed in the rate of learning and the parameters required across varying domains of applications including image classification and object detection.

In this paper, we approach the problem of keypoint localization as a semantic segmentation and tracking task unlike other approaches and propose a novel architecture that can leverage the spatio-temporal features in sequences to generate more consistent and robust detections. Using our proposed CoordConvGRU architecture, we demonstrate consistent performance improvement by conducting ablation studies on synthetic video sequences and compare our performance with other state-of-the-art methods in section \ref{Results}. We also make the network architectures and annotated data publicly available. \footnote{Project page and annotated KITTI sequences: \url{https://ayushgaud.github.io/Stacked_HG_CoordConvGRU/}}
\section{APPROACH}
\subsection{Overview}
Our approach builds on the notion that in a real-world context the data possess temporal consistency and the spatial features move gradually with time. Hence, instead of doing processing on static images we propose a network which can learn to model the motion dynamics in an end-to-end fashion. We leverage the recurrent neural network architecture which jointly learns to model spatial and temporal features from videos.
\subsection{Network Architecture}
\begin{figure}[ht!]
    \begin{center}
    \includegraphics[width=8cm, height=3.5cm] {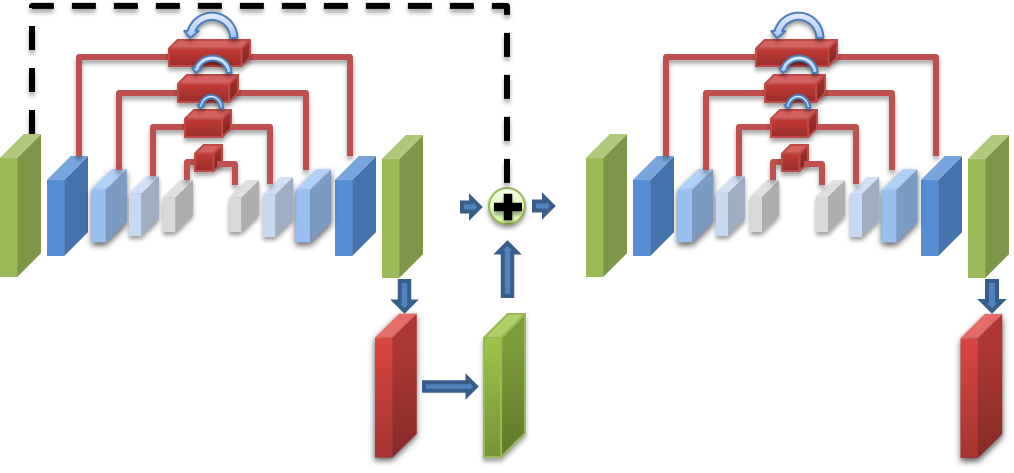}
    \caption{Two Stack Hourglass with Recurrent Cells for Skip Layers}
    \label{fig:architecture}
    \end{center}
\end{figure}
\begin{figure}
    \begin{center}
    \includegraphics[width=8cm, height=5.2cm] {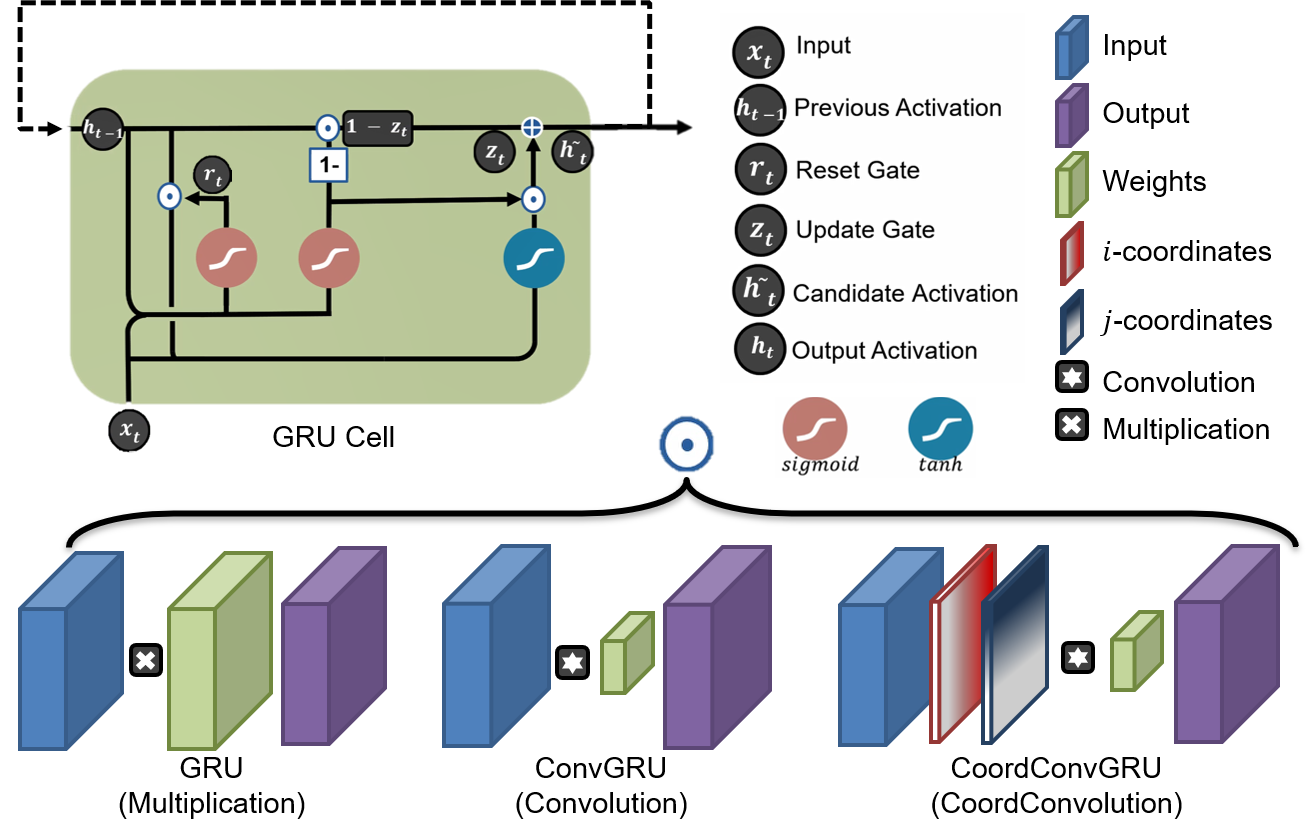}
    \caption{Comparison of different types of GRU cells}
    \label{fig:gru_cells}
    \end{center}
\end{figure}
\subsubsection{Stacked Hourglass with Intermediate Supervision}
Stacked hourglass networks proposed by Newell et al.\cite{newell2016stacked} shows that stacking conserves the higher order spatial relationships. Stages of hourglass enable inferring higher order features when intermediate supervision is applied. We utilize a similar architecture with iterative refinement using multiple stacks of hourglass for predicting keypoint likelihood map.

\subsubsection{Convolution Gated Recurrent Units (ConvGRU)}
Conventional Recurrent Neural Networks are applied to sequence of inputs to capture the temporal relation in the data. However, due to vanishing gradient problem gated architectures were proposed. Long Short Term Memory (LSTM), one of the most popular model used for RNNs has three gates namely, input, output and forget. The latter controls the amount of information flow from the previous states and acts as a memory for predictions. Gated Recurrent Unit (GRU) acts just like LSTMs but with a much efficient architecture by assuming a correlation between memorizing and forgetting. This enables it to use only a single gate to control the two and consequently the output flow. However, for using RNNs with images, they are vectorized into large 1D arrays which increases the number of parameters to learn and loses the spatial connectivity between pixels. This makes the network harder to train due to large search space where spatial context is already lost. To mitigate this problem convolutional recurrent units were introduced which, unlike regular GRU replaces dot products with small convolution filters as shown in the equation below
\begin{subequations}
\begin{equation}
    z _ { t } = \sigma \left( W _ { h z } * h _ { t - 1 } + W _ { x z } * x _ { t } + b _ { z } \right)
\end{equation}
\begin{equation}
    r _ { t } = \sigma \left( W _ { h r } * h _ { t - 1 } + W _ { x r } * x _ { t } + b _ { r } \right)
\end{equation}
\begin{equation}
    \hat { h _ { t } } = \Phi \left( W _ { h } * \left( r _ { t } \odot h _ { t - 1 } \right) + W _ { x } * x _ { t } + b \right)
\end{equation}
\begin{equation}
    h _ { t } = \left( 1 - z _ { t } \right) \odot h _ { t - 1 } + z \odot \hat { h } _ { t }
\end{equation}
\end{subequations}
Learning such small filters instead of weights for each pixel proves to be much more efficient while at the same time preserving spatial context.
\begin{figure*}[ht!]
    \begin{center}
    \includegraphics[width=17cm, height=7cm] {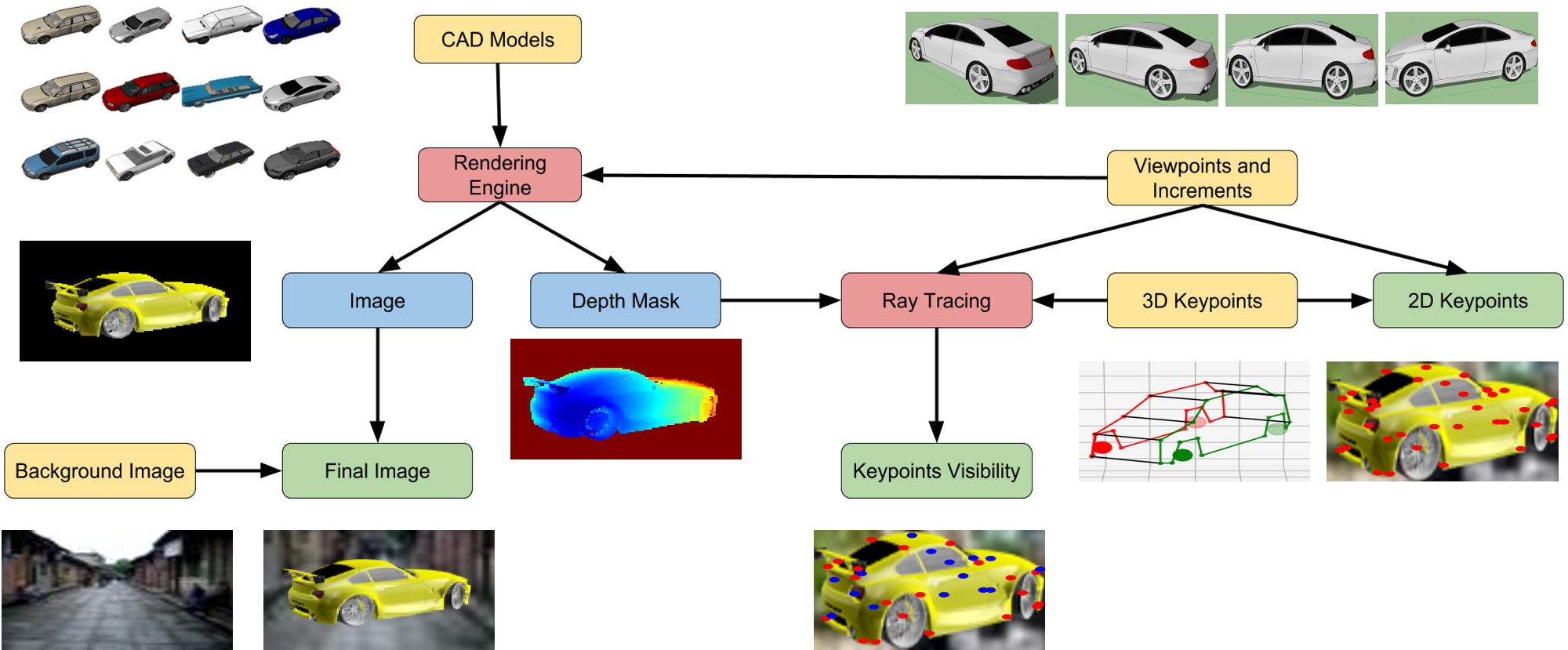}
    \caption{Synthesis Pipeline: The above figure depicts the synthetic dataset generation flow. ShapeNet\cite{chang2015shapenet} CAD models are rendered with Blender from various viewpoints in a smooth motion for generating sequences. Green blocks represent the output of the pipeline which includes the sequence images with background, its corresponding 2D keypoints and a list of visible keypoints.}
    \label{fig:synthesis_pipeline}
    \end{center}
\end{figure*}
\subsubsection{CoordConv Layer}
CoordConv layers\cite{liu2018intriguing} are a simple extension to regular convolution layers. They add two extra channels (in case of images) filled with coordinate information which are concatenated at the end of a convolution layer. This enables the network to flexibly learn translational invariance. It means that zero weights for these channels would imply full translational invariance like standard convolution while any other value would lead the network to learn translation invariance upto a degree which becomes a training parameter depending on the task.
\subsubsection{Proposed Network Architecture}
We introduce an improved architecture based on the stacked hourglass design for preserving both spatial and temporal context in a sequence of images. Capitalizing on the above notions, we present a recurrent architecture by replacing standard convolutions for skip connections with ConvGRU layers. The RNN cells are stacked together for processing a sequence of four images, while the weights among the stacks for each frame is shared within the graph. Since in the original paper\cite{newell2016stacked} there was only a marginal performance improvement observed when comparing eight stack hourglass network with a two stack network, hence we also consider a two stack architecture with intermediate supervision. Each stack consists of five layers of convolution filters with max pooling operation with pool size of (2,2). The output of each layer is passed to the next convolution layer and the skip layer consisting of GRU cell. This is followed by a series of upsampling layers and subsequent convolutions to bring the output size to 64x64. Each convolution filter has a channel size of 36 corresponding to each keypoint heatmap except in case of CoordConvGRU cells where two channels are added for each $i$ and $j$ coordinates as shown in figure \ref{fig:gru_cells}. We use four residual modules each embedded with an RNN cell for both stacks. We train the network on 64x64 images. The network outputs 36 heatmaps of size 64x64 at the end of every stack along with the states of the recurrent cells. We further show that our proposed cell which is created by modifying the ConvGRU layers, and using CoordConv layers i.e adding two channels one for $X$ and one for $Y$ axis, leads to further improvement in performance. The implementation of this architecture is publicly made available and can be downloaded from here.\footnote{Tensorflow implementation: \url{https://github.com/ayushgaud/hourglass_CoordConvGRU}}

\subsection{Loss Function}
Newell et al.\cite{newell2016stacked} in the original stacked hourglass paper treats the keypoint localization task as a regression problem and hence uses mean squared error loss for optimizing over the heatmaps. We posit a different route by treating this task as a classification problem. This notion stems from the similarity between the semantic segmentation and keypoint localization task. Due to this, we use sigmoid cross entropy as a loss function over heatmaps. These heatmaps are computed by using Gaussian distribution centered at ground truth location of keypoints with standard deviation of 1 pixel. In our case we represent the model of vehicle using 36 keypoints annotated at semantically meaningful joint locations. If predicted heatmaps for $i^{th}$ image in sequence of length $N$ is represented by $X_i$ and labeled heatmap is represented by $Z_i$. Then the sigmoid cross entropy loss can be written as:
\begin{equation}
loss = \sum _ {i = 1} ^ {N} X_i - X_i * Z_i + log(1 + e ^{-X_i})
\end{equation}
We minimize this loss between predicted heatmaps from both stacks and ground truth heatmaps leveraging the benefits of intermediate supervision along with standard form of supervision done at the end of the network.
\subsection{Dataset Generation and Training}
\subsubsection{Synthetic Dataset Generation}
Since we require large amount of data to train our network and annotated sequences are hard to find, We use the 3D keypoint annotations from \cite{li2016deep} and render CAD models of cars from ShapeNet\cite{chang2015shapenet} using the Render for CNN \cite{Su_2015_ICCV} pipeline. To enable sufficient variations in the data we randomly sample initial viewpoints and their respective increments using a uniform distribution. These viewpoint increments helps in generating desirable smooth realistic motions. We also vary the number of light sources and their intensity randomly for each sequence of data. An overview of the  synthesis pipeline is given in figure \ref{fig:synthesis_pipeline}. We use 472 3D CAD models and their corresponding annotations provided by \cite{zia2015towards}. The 36 semantically meaningful joints on each car are projected from the viewpoints at which the images are rendered. To identify visible set of keypoints for evaluation, we also render depth channel data and record camera viewpoints for each frame. We then use this information and perform ray tracing to identify the depths of each keypoint. This depth is compared to the depth of the projected keypoints from the rendered depth image and the mismatching keypoints are identified to be occluded. Each synthetic image is cropped and overlayed on a realistic background image to avoid overfitting. To maintain consistency the background image is kept same across the frames of each sequence. A total of hundred sequences with four frames each are generated for every CAD model.  
\subsubsection{Training Details}
The proposed network is built on the TensorFlow framework. We initialize the weights using Xavier initialization scheme with a learning rate of $2.5*10^{-4}$. We use RMSProp optimizer and decay the learning rate exponentially at every 20000 steps by a factor of 0.96. The input to the network are images of size is 64x64 at batches of 60 for initial training. Once the loss plateaus, the backpropagation is performed on the unwrapped recurrent network with shared pre-trained weights on a sequence length of 4 frames and a batch size of 16. The loss is computed on all the frames in the sequence and used for optimizing the network. We stop the training at around 500K iterations in total which takes almost 2 days to complete on 4 Nvidia 1080Ti GPUs.

\section{RESULTS}\label{Results}
\subsection{Keypoint Localization}
\begin{figure}
    \begin{center}
    \includegraphics[width=8cm, height=3.5cm] {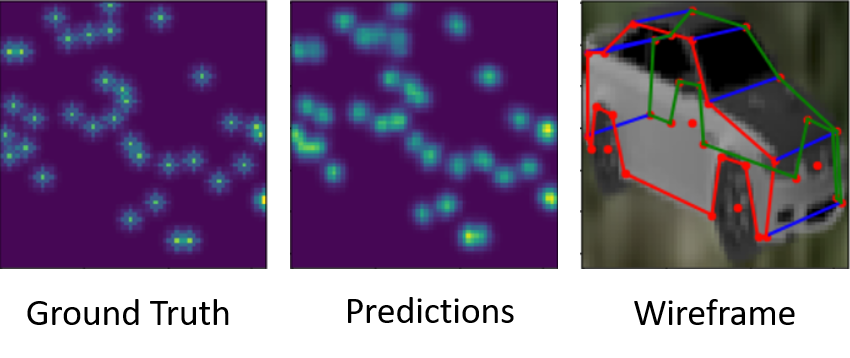}
    \caption{Ground truth and predicted keypoint likelihood maps (heatmaps)}
    \label{fig:predictions}
    \end{center}
\end{figure}

We evaluate the performance of our network using the PCK metric. This is a standard method used to label a keypoint detection as correct depending on whether or not it lies within the circle of radius $\alpha*L$ centered at the ground truth keypoint location where $L$ represents the larger dimension of the image. Instead of using the standard $\alpha=0.1$ we use a much stricter threshold of 0.05. Even though the network is trained on all the keypoints irrespective of visibility, we generate quantitative results based on the visible set of keypoints to remain consistent with the convention. It should be noted that the wireframes are made by joining all the predicted keypoint locations, including the occluded ones in a structural fashion. Figure \ref{fig:predictions} shows the ground truth and the predicted heatmap from the network. We take argmax over each of the 36 predicted heatmaps of all the joints and construct a wireframe by joining these points.
% \begin{figure}
%     \begin{center}
%     \includegraphics[width=8cm, height=6cm] {figures/synthetic.png}
%     \caption{Qualitative predictions on synthetically rendered images}
%     \label{fig:qual_synth}
%     \end{center}
% \end{figure}
\subsection{Ablation Experiments}
Given that our main contributions are based on the network architecture, we explore the effect of adding recurrent cells by comparing the 2D keypoint localization accuracy. To make a fair comparison we train all the networks from scratch on the same rendered synthetic dataset considering standard hourglass network as a baseline. We train three different network architectures including the baseline hourglass and tabulate the PCK measure at $\alpha=0.05$ in Table \ref{table:synt_quant}. 

It is interesting to note that adding ConvGRU layers itself improves the mean 2D keypoint accuracy from 94.8\% to 96.1\% thereby proving to have learnt to model the motion dynamics between the frames in sequences. We also observe that further changing the RNN cells in residual layers from ConvGRU to CoordConvGRU improved the result by another 0.6\% at 96.7\%. This improvement could be attributed to the selective translational invariance nature of the layer which enables it to capture the spatio-temporal consistencies in the sequence.
\begin{table}[h!]
\centering
\begin{tabular}{ |c|C{2cm}|}
\hline
Network & PCK @ $\alpha = 0.05$ \\
\hline
Standard Hourglass (Baseline) & 94.81 \\
\hline
Hourglass ConvGRU + Sequence & 96.10 \\
\hline
Hourglass CoordConvGRU + Sequence & 96.71 \\
\hline
\end{tabular}
\caption{PCK Comparison results between the baseline hourglass network and proposed network architectures.}
\label{table:synt_quant}
\end{table}

\begin{table}[h!]
\centering
\begin{tabular}{ |c|C{2cm}|}
\hline
Network & PCK @ $\alpha=0.1$ \\
\hline
DDN \cite{yu2016deep} & 67.6 \\
\hline
WN-gt-yaw \cite{kanazawa2016warpnet} & 88.0 \\
\hline
Zia et al. \cite{zia2015towards} & 73.6 \\
\hline
DSN-2D & 27.2 \\
\hline
DSN-3D & 76.0 \\
\hline
plain-2D & 45.2 \\
\hline
plain-3D & 88.4 \\
\hline
DISCO-3D-2D & 90.1 \\
\hline
DISCO-vis-3D-2D & 92.3 \\
\hline
DISCO-Vgg & 83.5 \\
\hline
DISCO \cite{li2017deep} & \textbf{93.1} \\
\hline
Standard Hourglass &  82.11\\
\hline
Hourglass CoordConvGRU & 88.27 \\
\hline
\end{tabular}
\caption{PCK accuracies on KITTI 3D dataset\cite{zia2015towards}}
\label{table:kitti_3d}
\end{table}
\begin{table}[h!]
\centering
\begin{tabular}{ |c|C{2cm}|}
\hline
Network & PCK @ $\alpha = 0.1$ \\
\hline
Standard Hourglass & 82.33 \\
\hline
Hourglass CoordConvGRU & 87.25 \\
\hline
Hourglass CoordConvGRU + Sequence & 87.81 \\
\hline
\end{tabular}
\caption{PCK comparison on annotated KITTI sequences}
\label{table:kitti_seq}
\end{table}
\begin{figure*}[ht!]
    \begin{center}
    \includegraphics[width=16cm, height=8cm] {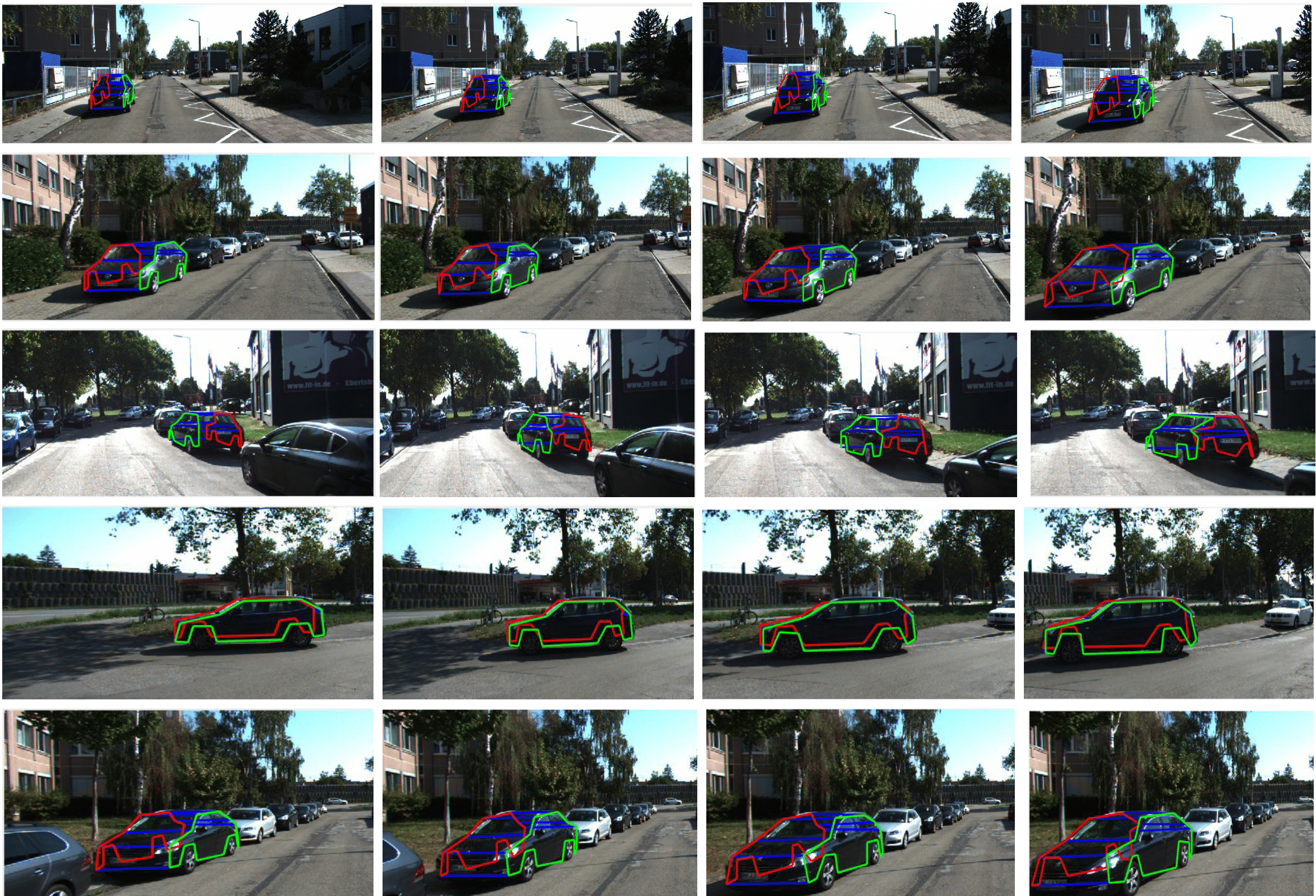}
    \caption{Wireframes predicted by the network on KITTI image sequences}
    \label{fig:real}
    \end{center}
\end{figure*}
\subsection{Results on KITTI images}
We fine tune the proposed architectures on 300 real images from KITTI. The networks are not refined on sequential data but only static images. The qualitative results in figure \ref{fig:real} shows that our networks are able to handle this data distribution transform from synthetic to real quite effectively with minimal training. We also perform a quantitative comparison of our approach on the annotated KITTI\cite{geiger2013vision} images. Table \ref{table:kitti_3d} reports PCK accuracies for various state-of-the-art methods including Zia et al.\cite{zia2015towards}, DNN\cite{yu2016deep}, WarpNet\cite{kanazawa2016warpnet} and various variants of DISCO\cite{li2017deep}. Since the DISCO architecture is based on supervision for intermediate tasks, we also report the comparison of PCK values for variants depending on the supervision performed. DISCO-vis-3D-2D, DISCO-
3D-2D, plain-3D and plain-2D are networks without pose, pose+visibility, pose+visibility+2D and pose+visibility+3D, respectively removed as a supervisory input to the complete deep supervision required for DISCO. It should be noted that we only provide 2D keypoint supervision for training and still achieve better results than most of the networks using a much simpler architecture. The amount of supervision required for training our network is equivalent to plan-2D and DSN-2D which perform significantly poorly on the keypoint localization task at PCKs 45.2\% and 27.2\% respectively. This shows that our proposed architecture at PCK 88.27\% has far better expressive capabilities compared to the other methods for this task. We also outperform standard hourglass by almost 6.1\% on static KITTI images from \cite{zia2015towards}.

We also use our labeled 524 sequences of images from KITTI for evaluation of performance of the network on sequences. In Table \ref{table:kitti_seq} we have tabulated the PCK values on this data. It is clear that again our proposed network surpasses our baseline by a significant margin of around 6\% and is 0.5\% higher than without sequence. It should also be noted that the disparity between performance is significant when compared to synthetic data. This can be because of the better convergence behavior of the CoordConv\cite{liu2018intriguing} unit of the RNN cells. Since we fine-tune the network on a limited set of images this behavior is not contrary to the expectations.

\subsection{Run-Time Analysis}
While using RNNs on sequential data gives significant performance improvements in terms of accuracy, they do come at a price. Run time of ConvGRU and CoordConvGRU is higher as they perform more operations than a regular convolution layer. At the same time since the CoordConv layers add two more channels to each convolution layer inside the RNN cell, this further increases the forward pass time of the network. Our network is much simpler in architecture than the state-of-the-art DISCO\cite{li2017deep} as it based on the very deep VGG network architecture with fully connected layers for intermediate task supervision. This enables us to run this network for applications requiring real-time time predictions. We evaluated the forward pass times of all the three networks and the results are presented in Table \ref{table:run_time}. Since while doing forward pass, the network only requires images and hidden states of the previous RRN cells hence the run time is computed for a single frame and not the sequence. 
\begin{table}[ht!]
\centering
\begin{tabular}{ |c|c|}
\hline
Network & Forward Pass Time \\
\hline
Standard Hourglass &  11ms\\
\hline
Hourglass ConvGRU &  19ms\\
\hline
Hourglass CoordConvGRU &  21ms\\
\hline
\end{tabular}
\caption{Run time of different networks on an Nvidia 1080Ti GPU}
\label{table:run_time}
\end{table}
\section{CONCLUSIONS}
We present a framework for keypoint localization on an object category and leverage the spatio-temporal correlation between frames in a sequence. The experimental results prove our hypothesis and demonstrate that our network performs at par with other networks which are trained at same level of supervision, using a relatively simpler architecture. We also demonstrate that our network is fast enough for real time applications like autonomous driving. We showcase a new type of RNN cell and conduct ablation experiments to highlight the performance benefits of using it. The results from the proposed architecture can further be improved by adding multiple levels of supervision and deserves further examination. We make our proposed network architecture and annotated data for KITII sequences publicly available for critical analysis and extension of this work.
\bibliographystyle{IEEEtran}
\bibliography{IEEE}
\end{document}